\documentclass{article}

\usepackage[preprint]{neurips_2026}

\usepackage[utf8]{inputenc} 
\usepackage[T1]{fontenc}    
\usepackage{hyperref}       
\usepackage{url}            
\usepackage{booktabs}       
\usepackage{amsfonts}       
\usepackage{nicefrac}       
\usepackage{microtype}      
\usepackage{xcolor}         
\setcitestyle{round}
\hypersetup{
    colorlinks=true,
    citecolor=blue,
}

\usepackage{amsmath}
\usepackage{graphicx} 
\usepackage{algorithm}
\usepackage{algorithmic}
\usepackage{multirow}
\usepackage{wrapfig}

\title{Orthrus: Memory-Efficient Parallel Token Generation via Dual-View Diffusion}

\author{%
  Chien Van Nguyen\thanks{Email: \texttt{chienn@uoregon.edu}} \\
  University of Oregon \\
  \And 
  Chaitra Hegde \\
  Google DeepMind
  \And
  Van Cuong Pham \\
  University of Oregon
  \AND
  Ryan A. Rossi \\
  Adobe Research
  \And
  Franck Dernoncourt \\
  Adobe Research
  \And
  Thien Huu Nguyen \\
  University of Oregon
}

\begin{document}

\maketitle

\begin{abstract}
We introduce {\bf Orthrus}, a simple and efficient dual-architecture framework that unifies the exact generation fidelity of autoregressive Large Language Models (LLMs) with the high-speed parallel token generation of diffusion models. The sequential nature of standard autoregressive decoding represents a fundamental bottleneck for high-throughput inference. While diffusion language models attempt to break this barrier via parallel generation, they suffer from significant performance degradation, high training costs, and a lack of rigorous convergence guarantees. Orthrus resolves this dichotomy natively. Designed to seamlessly integrate into existing Transformers, the framework augments a frozen LLM with a lightweight, trainable module to create a parallel diffusion view alongside the standard autoregressive view. In this unified system, both views attend to the exact same high-fidelity Key-Value (KV) cache; the autoregressive head executes context pre-filling to construct accurate KV representations, while the diffusion head executes parallel generation. By employing an exact consensus mechanism between the two views, Orthrus guarantees lossless inference, delivering up to a $7.8\times$ speedup with only an $O(1)$ memory cache overhead and minimal parameter additions. We release the code at \url{https://github.com/chiennv2000/orthrus}.
\end{abstract}

\section{Introduction}
Autoregressive (AR) Large Language Models (LLMs) are currently the predominant architecture in natural language processing, demonstrating robust performance across a diverse set of complex reasoning and generation tasks \citep{radford2019language,brown2020language,radford2018improving,touvron2023llama, achiam2023gpt, guo2025deepseek}. However, AR models suffer from a fundamental inefficiency during the decoding phase. While the pre-filling stage processes prompt tokens in parallel by leveraging self-attention, the generation phase computes tokens strictly sequentially. This one-by-one generation creates a memory-bandwidth bottleneck, leading to hardware underutilization and high inference latency.

Diffusion Language Models (DLMs) \citep{nie2025large, arriola2025block, zhu2025llada, ye2025dream}  natively bypass this bottleneck by generating blocks of tokens in parallel. Despite providing significant inference speedups, DLMs consistently underperform AR models of a similar scale and require massive training datasets to achieve baseline coherence. Recent approaches attempt to adapt pre-trained AR models into diffusion models to bridge this quality gap \citep{hu2024acdit, wu2025fast}. However, these adaptations remain computationally expensive, often requiring continuous pre-training up to 500B tokens, and still fail to match the exact predictive distribution of the original AR models due to architectural divergence.

To overcome this dichotomy, we propose resolving the trade-off at the fundamental architectural level by {\it unifying the strengths of both paradigms within a single Transformer}. We introduce Orthrus, a novel dual-architecture framework designed to natively support parallel generation without sacrificing the exact predictive distribution of the base autoregressive model. The core architectural insight of Orthrus is that the AR bottleneck is strictly confined to the generation phase; its self-attention mechanism remains optimal for building context representations. Consequently, Orthrus freezes the pre-trained AR model and utilizes its standard forward pass exclusively during the pre-filling stage to compute a high-fidelity Key-Value (KV) cache. To enable high-speed parallel generation, we structurally augment the network by integrating a lightweight, trainable diffusion module directly alongside the AR attention heads.

This structural unification allows both views to operate over the exact same context, inherently resulting in zero redundant cache overhead. During generation, the diffusion head conditions directly on the high-quality KV cache constructed by the AR head to generate multiple future tokens in parallel. To strictly guarantee lossless inference, the framework incorporates an intrinsic two-head consensus mechanism: token trajectories generated by the diffusion view are structurally validated by the frozen AR view, guaranteeing that the final output strictly matches the base model's exact predictive distribution. By decoupling the parallel generation mechanism from the sequential constraints of the base model, Orthrus achieves exact inference parity at significantly accelerated speeds.

In summary, our main contributions are:
\begin{itemize}
    \item {\bf A Novel Dual-Architecture Framework:} We introduce Orthrus, a structural unification that embeds a parallel diffusion module within a standard AR Transformer, allowing both views to operate over a shared KV cache with {\bf zero redundant historical KV cache storage.} Using intra-model consensus, it preserves the exact predictive distribution of the base LLM, ensuring strictly lossless generation that outperforms prior diffusion adaptations.
    \item {\bf Significant Inference Acceleration:} By natively exploiting the diffusion head for parallel token generation, Orthrus successfully breaks the sequential bottleneck, delivering up to a $7.8\times$ speedup.
    \item {\bf Extreme Parameter and Memory Efficiency:} The architectural integration is highly lightweight. Parallel capabilities can be injected into strong AR baselines by fine-tuning only 16\% of the total model parameters using less than 1B tokens (requiring under 24 hours on a single 8xH200 node).
    
\end{itemize}

\section{Preliminaries}
\label{sec:preliminaries}

To contextualize the architectural design of our proposed framework, we formalize the distinct probability modeling paradigms of Autoregressive (AR) and Masked Diffusion Language Models (MDMs). This formulation isolates the mathematical trade-off between generation quality and inference speed, establishing the foundation for our structural unification.

\subsection{Autoregressive and Diffusion Paradigms}

\paragraph{Autoregressive Language Modeling.} AR models learn the true data distribution by factorizing the joint probability of a sequence $\mathbf{x} = (x_1, x_2, \dots, x_N)$ using the exact chain rule of probability $p_{\text{AR}}(\mathbf{x}) = \prod_{i=1}^{N} p_{\theta}(x_i \mid \mathbf{x}_{<i})$. The model parameters $\theta$ are typically optimized via the negative log-likelihood over the data distribution $\mathcal{D}$:
\begin{equation} \label{eq:ar_nll}
    \mathcal{L}_{\text{AR}}(\theta) = -\mathbb{E}_{\mathbf{x} \sim \mathcal{D}} \left[ \sum_{i=1}^N \log p_{\theta}(x_i \mid \mathbf{x}_{<i}) \right]
\end{equation}
By imposing no conditional independence assumptions, this formulation ensures each token $x_i$ is strictly conditioned on the entire preceding trajectory. While this causal dependency achieves state-of-the-art fidelity, it mandates sequential sampling. During inference, generating $N$ tokens requires $N$ distinct forward passes, repeatedly loading the Key-Value (KV) cache creating a fundamental, memory-bandwidth-bound bottleneck \citep{leviathan2022fast,adnan2024keyformer,ho2024block}.

\paragraph{Masked Diffusion Language Models.} Diffusion Language Models (DLMs) bypass the sequential bottleneck by framing generation as a parallel denoising process. Given a historical context $\mathbf{c} = \mathbf{x}_{\le t}$ and a corrupted block of future tokens $\mathbf{y}^t$, the reverse process trains a network parameterized by $\phi$ to predict the original tokens $\mathbf{y}^0$ simultaneously:
\begin{equation} \label{eq:mdm_loss}
    \mathcal{L}_{\text{MDM}}(\phi) = -\mathbb{E}_{\mathbf{x} \sim \mathcal{D}, t, \mathbf{y}^t} \left[ \sum_{k \in \mathcal{M}} \log p_{\phi}(y_k^0 \mid \mathbf{c}, \mathbf{y}^t) \right]
\end{equation}
where $\mathcal{M}$ is the set of masked indices. For highly accelerated inference (where denoising steps $T \ll |\mathcal{M}|$), the model relies on a strong conditional independence assumption:
\begin{equation} \label{eq:dlm_approx}
    p_{\text{DLM}}(\mathbf{y}^0 \mid \mathbf{c}) \approx \prod_{k \in \mathcal{M}} p_{\phi}(y_k^0 \mid \mathbf{c}, \mathbf{y}^t)
\end{equation}
While this formulation heavily amortizes memory-bandwidth costs by computing the entire block in a single forward pass, it inherently violates the strict causal dependency of the autoregressive model. Because the prediction of token $y_k$ does not condition on the exact, realized token $y_{k-1}$, the joint probability distribution modeled by the DLM drifts from the true AR target distribution \citep{ma2025dkv, chen2025dparallel, wu2025fast}.

\subsection{The Limits of Adaptation and Structural Unification}

To mitigate the high computational costs of training DLMs from scratch, recent works explore adapting pre-trained AR models into diffusion frameworks \citep{tian2025next,gat2025set,wu2025fast,cheng2510sdar,zhou2026dllm}. These approaches repurpose the robust representations of AR baselines by fine-tuning them on block-wise masked diffusion objectives (Equation \ref{eq:mdm_loss}). While these methods transition the model from sequential to parallel generation, adaptation fundamentally alters the base model, introducing severe performance trade-offs. This distributional drift is particularly catastrophic for reasoning-heavy tasks: during long-horizon generation, conditional errors compound rapidly, causing severe performance degradation. For instance, state-of-the-art adaptations like Fast-dLLM-v2 \citep{wu2025fast} suffers an 11-point accuracy drop on MATH-500 \citep{hendrycks2020measuring} relative to its AR baseline. Furthermore, because these adapted models typically rely on multiple iterative filtering steps during inference to recover coherence, they often negate the theoretical speed advantages of parallel decoding, resulting in marginal latency improvements. By modifying the base weights and discarding the strict sequential forward pass, adapted models lose the ability to recover the exact predictive distribution of the original baseline, cementing the structural trade-off between speed and fidelity.

The mathematical dichotomy establishes that exact causal conditioning ensures high fidelity but forces sequential computation, while conditional independence (Eq.~\ref{eq:dlm_approx}) enables parallelism at the cost of distributional drift. We resolve this tension by structurally unifying both paradigms at the attention level. Rather than permanently converting the base model, Orthrus decouples parallel generation from sequential constraints by grounding it within the frozen, high-fidelity representations of the AR baseline. We detail this dual-architecture design in Section~\ref{sec:methodology}.

\section{Methodology: The Orthrus Architecture}
\label{sec:methodology}

The design of Orthrus is rooted in a fundamental architectural trade-off: standard autoregressive (AR) models produce high-fidelity representations due to their strict causal conditioning, yet are bottlenecked by sequential generation. Conversely, parallel diffusion generation offers rapid decoding but often suffer from conditional drift and lower representation quality. To reconcile this trade-off, Orthrus introduces a unified dual-view architecture. By injecting a lightweight diffusion head into a pre-trained AR model, we preserve its exact representation space while enforcing a strict functional decoupling: the frozen AR head is dedicated exclusively to constructing high-fidelity context representations, and the trainable diffusion head is specialized for high-speed parallel generation.

\subsection{Unified Dual-View Attention Mechanism}
Consider a prompt sequence $\mathbf{x}_{1:t} = (x_1, \dots, x_t)$. During prefilling, the frozen AR backbone $\mathcal{M}_{\text{AR}}$ processes the full context in \textbf{a single forward pass}, producing causal Key-Value representations $(\mathbf{K}_{\text{AR}}, \mathbf{V}_{\text{AR}})$. At generation time, however, producing $K$ continuation tokens requires \textbf{$K$ sequential forward passes}, each conditioned on all prior KV states, a fundamental memory-bandwidth bottleneck that our architecture is designed to eliminate.

\begin{figure}[t]
\centering
\includegraphics[width=0.9\linewidth]{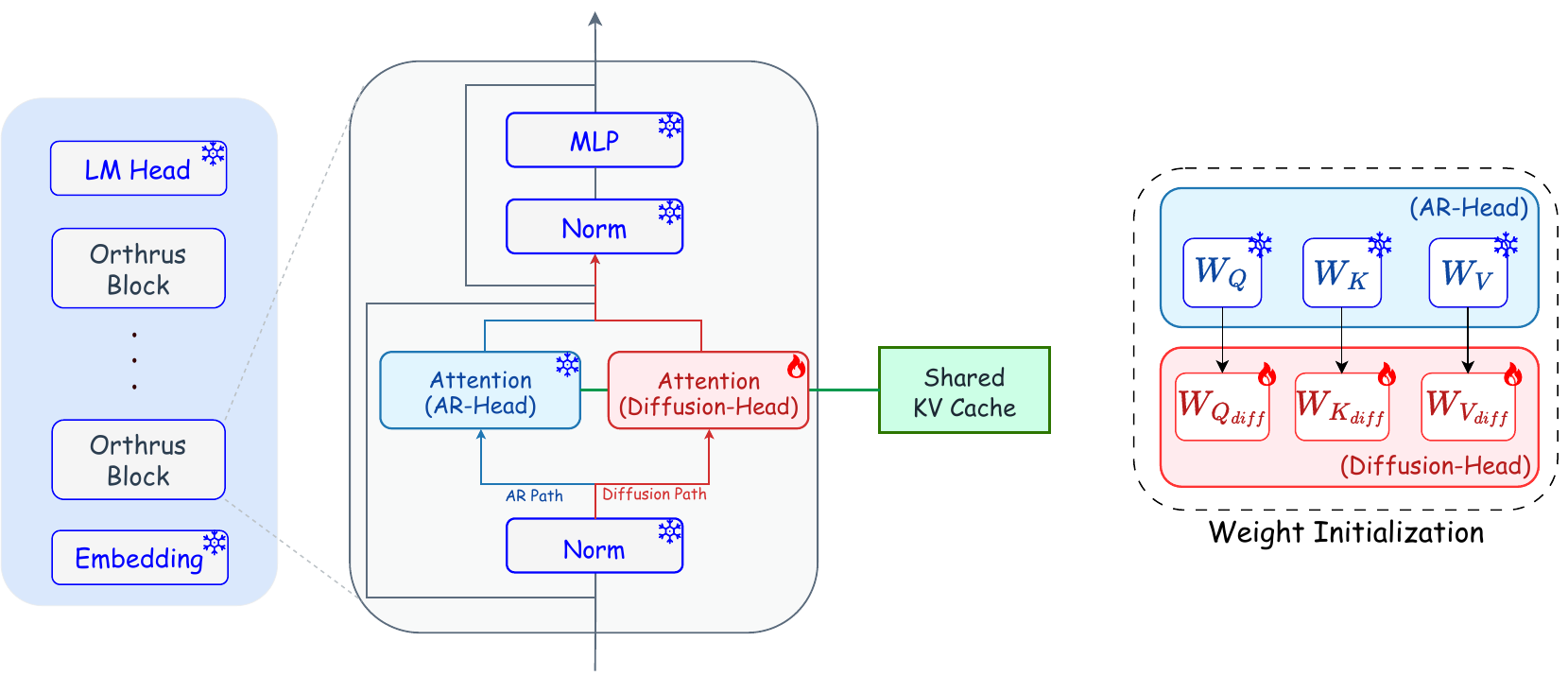}
\caption{\textbf{The Orthrus dual-view architecture.} Each Orthrus block features two distinct, parallel attention paths: a frozen AR head (blue) and a trainable diffusion head (red). The frozen AR head is used to encode context into KV representations, while the diffusion head enables parallel token generation. Both paths seamlessly attend over this single shared cache.}
\label{fig:orthrus_diagram}
\end{figure}

\paragraph{Parallel Diffusion View.}
We augment each transformer layer with a trainable diffusion attention module, parameterized by projection matrices $(\mathbf{W}_{Q_{\text{diff}}}, \mathbf{W}_{K_{\text{diff}}}, \mathbf{W}_{V_{\text{diff}}})$ initialized from their frozen AR counterparts, as illustrated in Figure~\ref{fig:orthrus_diagram}. To generate $K$ tokens in a single forward pass, we construct an extended sequence by concatenating the first token decoded by the AR view with $K{-}1$ \texttt{<mask>} embeddings, forming a parallel block of $K$ positions. These positions are processed simultaneously through the diffusion view, whose queries attend jointly over the frozen AR cache and the bidirectional self-representations of the mask block:
\begin{equation} \label{eq:orthrus_attn}
    \mathbf{O}_{\text{diff}} = \text{Softmax}\!\left(\frac{\mathbf{Q}_{\text{diff}}\,[\mathbf{K}_{\text{AR}} \,\|\, \mathbf{K}_{\text{diff}}]^{\top}}{\sqrt{d_{\text{head}}}}\right) [\mathbf{V}_{\text{AR}} \,\|\, \mathbf{V}_{\text{diff}}],
\end{equation}
where $[\cdot\|\cdot]$ denotes concatenation along the sequence axis and $\mathbf{O}_{\text{diff}} \in \mathbb{R}^{K \times d_{\text{head}}}$ contains the hidden states for all $K$ parallel positions. Two structural properties follow directly. Because $(\mathbf{K}_{\text{AR}}, \mathbf{V}_{\text{AR}})$ are reused in-place from the prefill pass, so the diffusion view introduces \textbf{zero additional historical KV cache memory}. Since only $(\mathbf{W}_{Q_{\text{diff}}}, \mathbf{W}_{K_{\text{diff}}}, \mathbf{W}_{V_{\text{diff}}})$ are updated during training, the total number of trainable parameters is approximately $16\%$ of the full model.

\begin{figure}[t]
\centering
\includegraphics[width=0.9\linewidth]{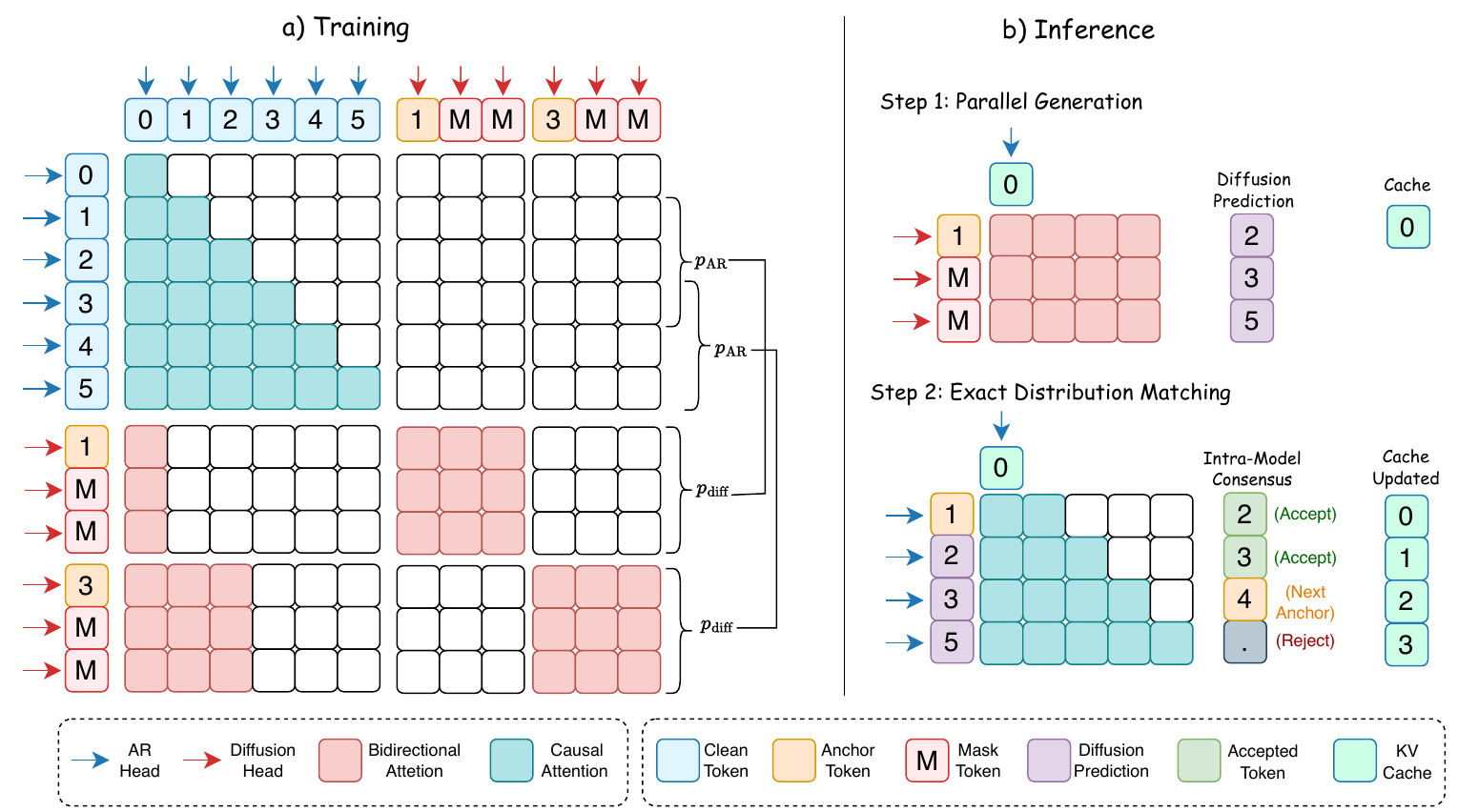}
\caption{\textbf{The Orthrus dual-view attention mechanism.} \textbf{(a) Training:} The AR path (blue arrows) processes the \textbf{clean context} using standard causal masking to establish the exact target distribution. The diffusion path (red arrows) processes \textbf{corrupted parallel blocks} (an anchor plus \texttt{<mask>} tokens). The diffusion head attends directly to the KV representations constructed by the AR path, and its parallel predictions ($p_{\text{diff}}$) are distilled to match the exact corresponding AR rows ($p_{\text{AR}}$). \textbf{(b) Inference:} The diffusion head projects $K$ candidate tokens in parallel (Step 1), which the AR head validates in a single pass (Step 2). Accepted tokens' KV states are seamlessly appended to the shared cache.}
\label{fig:orthrus_mechanics}
\end{figure}

\subsection{Training: Dual-Pass Block Masking}

Because the AR backbone is strictly frozen, training reduces to aligning the diffusion view's parallel predictions with the AR model's exact target distribution. Given a sequence $\mathbf{x} = (x_1, \dots, x_L)$, we sample $B$ random anchor positions $\{a_b\}_{b=1}^{B}$ and extract contiguous blocks of length $K$, forming clean blocks $\mathbf{y}_b = (x_{a_b}, \dots, x_{a_b+K-1})$. Each block is corrupted by retaining the first token as a visible anchor and replacing the remaining $K{-}1$ positions with \texttt{<mask>} tokens:
\begin{equation}
    \tilde{y}_{b,k} = \begin{cases} x_{a_b} & k = 1 \\ \texttt{<mask>} & k = 2, \dots, K \end{cases}
\end{equation}
The $B$ corrupted blocks are concatenated and processed against the frozen AR KV cache $(\mathbf{K}_{\text{AR}}, \mathbf{V}_{\text{AR}})$ computed over the full sequence.

\paragraph{Dual-pass attention mask for the diffusion view.}
While the frozen AR path processes the clean historical context utilizing standard causal masking (top rows of Figure~\ref{fig:orthrus_mechanics}(a), denoted by blue arrows), the trainable diffusion head processes the \textbf{corrupted parallel blocks} and requires a specialized routing mechanism to prevent data leakage. To enforce this correct information flow during training, we construct a structured block mask $\mathbf{M}_{\text{diff}}$ for the diffusion view (represented by the bottom rows and red arrows) implemented using FlexAttention \citep{dong2024flex}. For a diffusion query at position $q$ and a key at position $k$, attention is permitted if and only if:
\begin{equation}
    \mathbf{M}_{\text{diff}}[q, k] = \underbrace{\mathbf{1}[k < L] \cdot \mathbf{1}[k \leq a_b - 1]}_{\text{causal AR context}} \;\;|\;\; \underbrace{\mathbf{1}[k \geq L] \cdot \mathbf{1}\!\left[\lfloor q/K \rfloor = \lfloor (k-L)/K \rfloor\right]}_{\text{bidirectional within block}},
\end{equation}

This specialized mask enforces two disjoint viewing rules: (i) each position within the corrupted block attends causally to the clean AR context preceding its block anchor, preventing future leakage; and (ii) all positions within the same block attend bidirectionally to one another, enabling parallel context aggregation across the mask span. By explicitly mapping $\mathbf{M}_{\text{diff}}$ to the bottom rows of the attention matrix, this structural isolation ensures that the corrupted context, comprising the anchor token and subsequent \texttt{<mask>} tokens processed via the diffusion path (red arrows) can jointly predict the future trajectory without attending to other parallel blocks.

\paragraph{Training objective.}
During training, the diffusion view utilizes the \texttt{<mask>} tokens to predict the subsequent tokens within the block, minimizing the forward KL divergence against the full predictive distribution of the frozen AR model over all masked positions:
\begin{equation} \label{eq:kl_div}
    \mathcal{L}_{\text{Orthrus}} = \mathbb{E}_{\mathbf{x}, \{a_b\}} \left[ \sum_{b=1}^{B} \sum_{k=1}^{K} D_{\text{KL}}\!\left( p_{\text{AR}}(\cdot \mid \mathbf{x}_{\leq a_b + k - 1}) \;\|\; p_{\text{diff}}(\cdot \mid \mathbf{x}_{< a_b}, \tilde{\mathbf{y}}_b) \right) \right],
\end{equation}
where $p_{\text{AR}}(\cdot \mid \mathbf{x}_{\leq a_b + k - 1})$ is the full token distribution predicted by the frozen AR head at sequence position $a_b + k - 1$, and $p_{\text{diff}}(\cdot \mid \mathbf{x}_{< a_b}, \tilde{\mathbf{y}}_b)$ is the parallel prediction of the diffusion view at the corresponding masked position. This soft distillation objective transfers the full predictive distribution of the AR model into the diffusion view. Gradients flow exclusively through diffusion module and the AR backbone remains strictly frozen throughout.

\subsection{Inference: Exact Distribution Matching via Intra-Model Consensus}
\label{subsec:inference}

At inference time, the structural unification of Orthrus enables a continuous, high-throughput generation loop executed entirely over a singular KV cache. Let $\mathbf{x}_{\le t}$ denote the currently generated sequence prefix, and $(\mathbf{K}_{\text{AR}}^{<t}, \mathbf{V}_{\text{AR}}^{<t})$ its corresponding high-fidelity cache computed natively by the AR backbone. The Orthrus inference loop proceeds through a continuous cycle of projection and structural synchronization:

\paragraph{Parallel Block Projection.} 
To bypass the sequential bottleneck, the diffusion view utilizes the shared KV cache to project a continuous trajectory of future tokens. To initiate parallel generation, we construct a block $\tilde{\mathbf{y}}_t$ of size $K$ by taking the current anchor token $x_t$ and concatenating it with $K{-}1$ \texttt{<mask>} tokens. The diffusion head processes this entire extended block in {\bf a single parallel forward pass}. Unlike other DLMs that rely on multi-step iterative denoising, we empirically find that this single-step projection is substantially more efficient, achieving a strictly higher token-per-forward-pass ratio.
By conditioning directly on the high-fidelity KV cache $(\mathbf{K}_{\text{AR}}^{<t}, \mathbf{V}_{\text{AR}}^{<t})$ natively constructed by the AR view, this pass yields a full, simultaneous projection of $K$ candidate tokens $\hat{\mathbf{y}} = (\hat{y}_1, \dots, \hat{y}_K) \sim p_{\text{diff}}(\cdot \mid \mathbf{x}_{<t}, \tilde{\mathbf{y}}_t)$ (Figure~\ref{fig:orthrus_mechanics}(b), Step 1).

\paragraph{Intra-Model Distribution Matching.}
To guarantee that the parallel projection strictly recovers the target distribution without conditional drift, the trajectory $\hat{\mathbf{y}}$ must be mathematically aligned with the exact causal distribution of the base model. The architecture routes the fully materialized block $\hat{\mathbf{y}}$ through the frozen AR head. Because these $K$ positions are fully populated in the input sequence, the AR head computes the exact target probabilities $p_{\text{AR}}(v \mid \mathbf{x}_{\le t}, \hat{\mathbf{y}}_{1:k-1})$ for all $k \in \{1, \dots, K\}$ simultaneously in a \textbf{single forward pass}.

\paragraph{Architectural Consensus Mechanism.}
With both the parallel prior distribution $p_{\text{diff}}$ and the exact target distribution $p_{\text{AR}}$ computed within the same representational space, the architecture dynamically synchronizes the projected tokens via a strict left-to-right evaluation. The consensus mechanism enforces strict structural identity with the causal AR path. A projected token $\hat{y}_k$ is retained if and only if it matches the greedy AR prediction exactly:
\begin{equation} \label{eq:deterministic_alignment}
    \hat{y}_k = \arg\max_{v \in \mathcal{V}} p_{\text{AR}}(v \mid \mathbf{x}_{\le t}, \hat{\mathbf{y}}_{1:k-1})
\end{equation}
For diverse generation (with temperature $T>0$), the architecture leverages an exact rejection sampling to align the parallel projection with the target distribution, guaranteeing strictly lossless sampling \citep{leviathan2022fast}.
If structural divergence occurs at index $j \le K$, verification halts. The architecture commits the synchronized prefix $\hat{\mathbf{y}}_{1:j-1}$ alongside the exact causal correction token $y_j$ drawn directly from $p_{\text{AR}}$, and truncates the shared KV cache to step $t+j$ (Figure~\ref{fig:orthrus_mechanics}(b), Step 2). This synchronization preserves the exact predictive distribution of the base model, delivering strictly lossless inference acceleration.

\section{Experiments}
\label{sec:experiments}

\subsection{Experimental Setup}

\label{subsec:setup}
\paragraph{Baselines and Model Scalability.}
To demonstrate the scalability and generalizability of our dual-view architecture, we select the state-of-the-art Qwen3 model family \citep{yang2025qwen3} as our foundation baselines. Specifically, we evaluate the 1.7B, 4B, and 8B parameter variants to observe how Orthrus scales from small to standard large language models. The original autoregressive (AR) backbone of each model remains frozen, with only the injected diffusion attention module being optimized.

\paragraph{Evaluation Benchmarks.}
To rigorously test the capacity of the diffusion head to mirror exact causal distributions without conditional drift, we evaluate Orthrus across a diverse and highly complex suite of zero-shot reasoning and algorithmic tasks. For mathematical reasoning, we benchmark performance on GSM8K \citep{cobbe2021training}, MATH-500 \citep{hendrycks2020measuring}, and recent AIME challenges (AIME24, AIME25) \citep{aime-wiki}. For structural and programmatic generation, we utilize HumanEval \citep{chen2021evaluating}, MBPP \citep{austin2021program}, Pseudo2code \citep{ye2025longproc}, and LiveCodeBench-v5 \citep{jain2024livecodebench}. This comprehensive task selection ensures that our empirical claims are validated across long-horizon generative trajectories that strictly penalize distributional divergence.

\paragraph{Implementation Details.}
During training, we configure the parallel projection block size to $K=32$ across all model scales. 
To maximize throughput, we adopt a one-step prediction strategy for the masked block, which we find sufficient to produce high-quality for the diffusion prediction. The models are trained for two epochs on a dataset of 600K examples (detailed in Appendix~\ref{app:training_details}). For each training instance, we construct a clean text context with a maximum length of 2048 tokens and generate a corresponding corrupted sequence containing 256 masked blocks placed at random anchor positions. The autoregressive backbone remains strictly frozen, only the newly injected diffusion heads are updated. Training is conducted on a single 8×H200 GPU node, utilizing FlexAttention \citep{dong2025flexattention} with the FlashAttention-4 backend \citep{zadouri2026flashattention} to implement the customized training masks. Finally, to strictly evaluate the exact distributional alignment between the diffusion projections and the frozen AR teacher, all reported generation metrics and acceptance lengths rely on greedy decoding for deterministic evaluation. 

\subsection{Efficiency Benchmarking}
\label{subsec:efficiency}

\paragraph{Efficiency Metrics.} We isolate algorithmic efficiency using Effective Tokens Per Forward Pass:
\begin{equation}
    \text{TPF} = \frac{\text{Total Generated Tokens}}{\text{Total Forward Passes}}
\end{equation}

This hardware-agnostic metric quantifies the average token throughput per inference step. Relative speedups are benchmarked against autoregressive (AR) baselines, which are bounded to a maximum TPF of $1$. For Orthrus, each continuous generation cycle inherently requires exactly two forward passes. By guaranteeing at least one token per cycle, this establishes a strict theoretical lower bound of $0.5$ TPF ($1$ token per $2$ passes). However, by leveraging the parallel diffusion view to project token blocks in a single initial forward pass, Orthrus bypasses the sequential bottleneck of standard AR inference.

Furthermore, our architecture conceptually advances the goals of traditional speculative decoding. Unlike standard speculative paradigms that rely on external draft models, incurring significant memory overhead to maintain isolated KV caches, our intra-model approach achieves parallel acceleration natively over a single shared KV cache, making it highly optimal for high-throughput production. A discussion comparing our architecture against speculative drafting systems is detailed in Section~\ref{subsec:speculative}. 

\begin{table}[h]
\centering
\caption{\textbf{Efficiency Benchmarking.} We report Tokens Per Forward Pass (TPF) and relative speedup against the sequential AR baseline (which operates at {\bf TPF} = 1.0 and 1.0$\times$ speedup) for both greedy decoding (temperature $T=0$) and diverse sampling (temperature $T=1$).}
\label{tab:efficiency_summary}
\resizebox{\textwidth}{!}{%
\begin{tabular}{@{}lcccccccccccc@{}}
\toprule
\multirow{3}{*}{\textbf{Task}} & \multicolumn{4}{c}{\textbf{Qwen3-1.7B}} & \multicolumn{4}{c}{\textbf{Qwen3-4B}} & \multicolumn{4}{c}{\textbf{Qwen3-8B}} \\ 
\cmidrule(lr){2-5} \cmidrule(lr){6-9} \cmidrule(l){10-13} 
 & \multicolumn{2}{c}{\textbf{$T=0$}} & \multicolumn{2}{c}{\textbf{$T=1$}} & \multicolumn{2}{c}{\textbf{$T=0$}} & \multicolumn{2}{c}{\textbf{$T=1$}} & \multicolumn{2}{c}{\textbf{$T=0$}} & \multicolumn{2}{c}{\textbf{$T=1$}} \\
\cmidrule(lr){2-3} \cmidrule(lr){4-5} \cmidrule(lr){6-7} \cmidrule(lr){8-9} \cmidrule(lr){10-11} \cmidrule(l){12-13}
 & \textbf{TPF} & \textbf{Speedup} & \textbf{TPF} & \textbf{Speedup} & \textbf{TPF} & \textbf{Speedup} & \textbf{TPF} & \textbf{Speedup} & \textbf{TPF} & \textbf{Speedup} & \textbf{TPF} & \textbf{Speedup} \\ \midrule
GSM8K & 4.20 & 4.37$\times$ & 3.88 & 4.04$\times$ & 4.91 & 4.48$\times$ & 4.97 & 4.53$\times$ & 5.30 & 5.22$\times$ & 4.69 & 4.62$\times$ \\
MATH-500 & 4.71 & 4.74$\times$ & 4.47 & 4.50$\times$ & 6.02 & 5.76$\times$ & 5.62 & 5.38$\times$ & 6.35 & 5.95$\times$ & 5.80 & 5.43$\times$ \\
AIME24 & 4.33 & 5.65$\times$ & 3.84 & 5.01$\times$ & 5.00 & 6.06$\times$ & 4.70 & 5.70$\times$ & 5.63 & 6.81$\times$ & 4.7 & 5.68$\times$ \\
AIME25 & 3.89 & 4.80$\times$ & 3.55 & 4.38$\times$ & 5.17 & 5.86$\times$ & 4.37 & 4.95$\times$ & 5.25 & 6.03$\times$ & 4.40 & 5.05$\times$ \\
HumanEval & 2.75 & 3.07$\times$ & 2.60 & 2.90$\times$ & 3.57 & 3.67$\times$ & 3.37 & 3.46$\times$ & 3.94 & 3.68$\times$ & 3.37 & 3.15$\times$ \\
MBPP & 2.76 & 3.07$\times$ & 2.88 & 3.20$\times$ & 3.74 & 4.43$\times$ & 3.50 & 4.15$\times$ & 3.95 & 3.90$\times$ & 3.75 & 3.70$\times$ \\
Pseudo2code & 4.60 & 4.90$\times$ & 4.37 & 4.65$\times$ & 7.40 & 7.38$\times$ & 6.95 & 6.93$\times$ & 7.51 & {\bf 7.83}$\times$ & 7.29 & 7.60$\times$ \\
LiveCodeBench-v5 & 3.86 & 5.87$\times$ & 3.58 & 5.44$\times$ & 4.50 & 6.28$\times$ & 4.28 & 5.97$\times$ & 5.17 & 6.68$\times$ & 4.34 & 5.61$\times$ \\ \midrule
\textbf{Average} & \textbf{3.89} & \textbf{4.25$\times$} & \textbf{3.65} & \textbf{4.27$\times$} & \textbf{5.04} & \textbf{5.20$\times$} & \textbf{4.72} & \textbf{5.13$\times$} & \textbf{5.39} & \textbf{5.36$\times$} & \textbf{4.43} & \textbf{5.02$\times$} \\ \bottomrule
\end{tabular}%
}
\end{table}

Table~\ref{tab:efficiency_summary} details these efficiency gains across our evaluation suite. Orthrus delivers substantial inference acceleration on all reasoning and algorithmic tasks, achieving an average TPF of 5.39 at the 8B parameter scale. Crucially, unlike existing DLMs that inherently trade generation quality for inference speed, Orthrus {\it mathematically guarantees exact distributional parity with the AR baseline}, ensuring strictly lossless acceleration.

\subsection{Comparison with State-of-the-Art Diffusion Models}
\label{subsec:sota_comparison}

While diffusion language models offer a novel path to parallel decoding, they often suffer from significant conditional drift. Achieving baseline coherence in these models demands massive computational resources. For instance, adaptation approaches like SDAR \citep{cheng2510sdar} require continuous pre-training on 50B tokens, while models like Dream \citep{ye2025dream} are trained on upwards of 580B tokens. Despite these training costs, these models still exhibit performance degradation and struggle to math the high-fidelity reasoning capabilities inherent to autoregressive models.

Table \ref{tab:sota_comparison} contrasts Orthrus with state-of-the-art diffusion paradigms on complex mathematical and structural reasoning benchmarks. The results demonstrate a clear performance gap: existing diffusion-based models, including Dream \citep{ye2025dream}, Fast-dLLM-v2 \citep{wu2025fast}, LLaDA-1.5 \citep{zhu2025llada}, SDAR \citep{cheng2510sdar}, Mercury Coder \citep{khanna2025mercury} and Diffusion Gemini \citep{googledeepmind2025gemini}, consistently lag behind in accuracy. Crucially, even though SDAR is initialized from the same Qwen3 foundation model as our architecture, suffers from degraded performance. In contrast, Orthrus is computationally efficient. By decoupling the diffusion head from the frozen AR backbone, we avoid the destructive interference caused by full-model fine-tuning. We successfully inject parallel generation capabilities by fine-tuning only 16\% of the total model parameters on less than 1B tokens, a lightweight process requiring under 24 hours on a single 8xH200 node.

\begin{table}[h]
\centering
\caption{Performance comparison with SOTA Diffusion Models.}
\label{tab:sota_comparison}
\resizebox{\textwidth}{!}{%
\begin{tabular}{lccccccc}
\toprule
\textbf{Model} & \textbf{Params} & \textbf{GSM8K} & \textbf{MATH-500} & \textbf{AIME-24} & \textbf{AIME-25} & \textbf{HumanEval} & \textbf{MBPP} \\ \midrule
Dream-7B & 7B & 79.3 & 39.6 & - & - &   57.9 & 58.8 \\
LLaDA-1.5 & 7B & 82.4 & 57.4 & - & - &  52.4 & 42.8 \\
SDAR-Qwen3-8B & 8B & 91.7 & 78.6 & 10.0 & 10.0 & 78.7 & 72.0 \\
SDAR-Qwen3-30B-A3B & 30B & 91.4 & 77.8 & 16.7 & 10.8 & 71.6 & 87.2 \\
Fast-dLLM-v2 & 7B & 83.7 & 61.5 & - & - & 43.3 & 28.2 \\
Mercury Coder Small & - & - & - & - & - & 90.0 & 76.6 \\
Gemini Diffusion & - & - & - & - & 20.0 & 89.6 & 76.0 \\ \midrule
\textbf{Orthrus-Qwen3-8B} & \textbf{8.7B} & \textbf{96.0} & \textbf{86.2} & \textbf{28.3} & \textbf{23.3} & \textbf{95.1} & \textbf{93.4} \\ \bottomrule
\end{tabular}%
}
\end{table}

\begin{wrapfigure}{r}{0.5\textwidth}
    \centering
    \includegraphics[width=\linewidth]{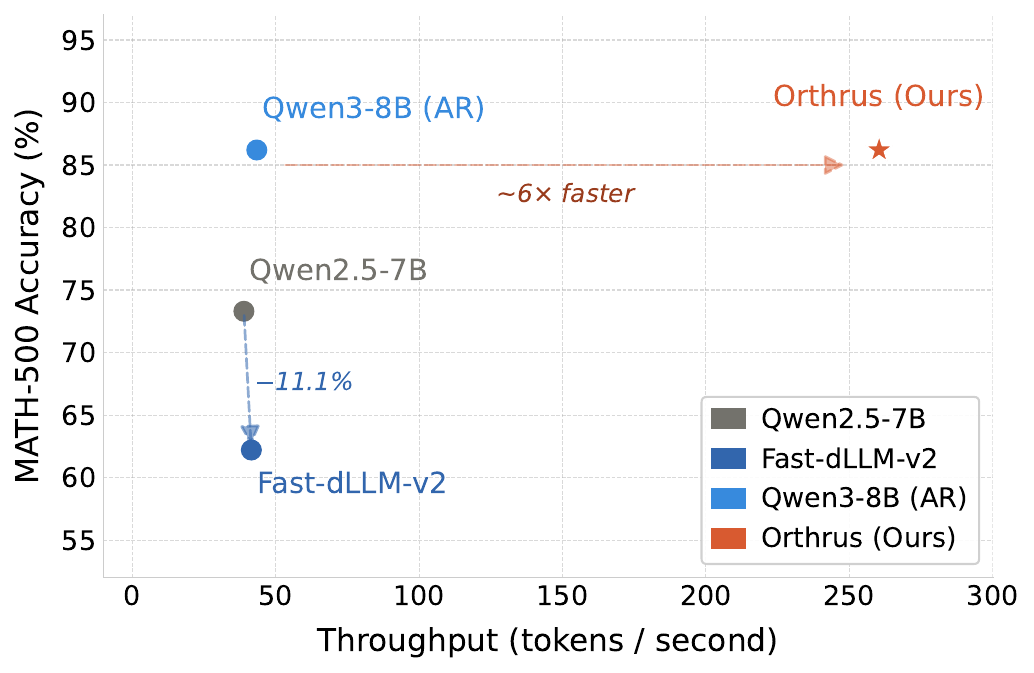}
    \caption{\textbf{Throughput vs. Accuracy on MATH-500.} Orthrus delivers a $6\times$ speedup over the Qwen3-8B baseline with strictly lossless performance, whereas Fast-dLLM-v2 suffers severe accuracy degradation.}
    \label{fig:math500_speed}
\end{wrapfigure}

Most importantly, because Orthrus relies on intra-model consensus rather than altering the base weights, its reasoning performance is directly inherited from, and upper-bounded by the selected frozen AR baseline. In our experiments, Orthrus achieves the exact zero-shot accuracy of the base Qwen3-8B model, establishing a new state-of-the-art for parallel generation fidelity. This structural property makes Orthrus a highly scalable, plug-and-play framework: {\it it can be seamlessly adapted to any high-quality existing open-source AR model} to unlock parallel throughput without sacrificing elite reasoning capabilities.

To further illustrate this structural advantage, Figure \ref{fig:math500_speed} visualizes the performance-efficiency trade-off in terms of absolute wall-clock throughput (tokens per second). Adaptation methods like Fast-dLLM-v2 incur a severe 11.1 point degradation on the MATH-500 benchmark relative to their AR baselines. Furthermore, their theoretical acceleration is often negated in practice, yielding negligible speedups due to the multiple iterative refinement steps required to recover output coherence. In contrast, Orthrus bypasses these inefficiencies entirely, delivering up to a 6$\times$ speedup with strictly lossless generation.

\subsection{Comparison with Speculative Decoding}
\label{subsec:speculative}

We contextualize Orthrus against state-of-the-art speculative decoding paradigms, specifically EAGLE-3 \citep{li2025eagle} and DFlash \citep{chen2026dflash}, evaluated on the Qwen3-8B foundation model. Standard speculative frameworks rely on training a distinct drafter model to rapidly project candidate tokens, which the larger base model subsequently verifies. While this decoupled approach mitigates sequential latency, it introduces a severe memory bottleneck: the system must maintain isolated, redundant KV caches for both the drafter and the verifier during inference. In contrast, Orthrus presents a structurally unified alternative. Because our parallel diffusion head conditions on the exact same KV representation space as the autoregressive (AR) backbone, it eliminates the need for an external drafter. This intra-model approach achieves parallel acceleration natively, resulting in zero redundant cache overhead (a detailed empirical analysis of memory consumption is provided in Appendix~\ref{app:analysis}).

\begin{figure}[htbp]
    \centering
    \includegraphics[width=\textwidth]{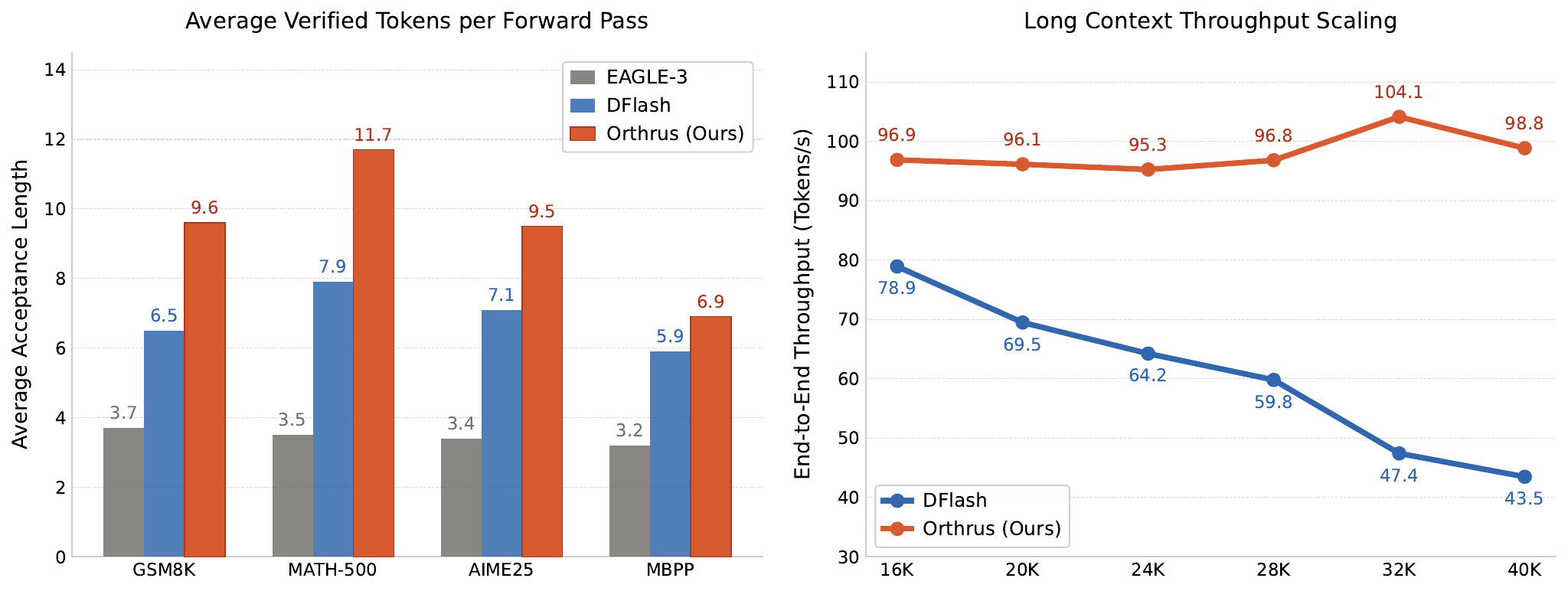} 
    \caption{\textbf{Performance scaling across benchmarks and context lengths.} 
    \textit{Left:} Average verified tokens per forward pass compared to state-of-the-art speculative decoding methods. 
    \textit{Right:} End-to-end throughput across scaling context lengths. Orthrus sustains robust generation speeds up to 40K tokens, avoiding the severe performance degradation exhibited by traditional speculative decoding (DFlash).}
    \label{fig:performance_combined}
\end{figure}

To evaluate this efficiency advantage, we analyze performance across two key dimensions: acceptance rate and long-context stability. First, we compare the Average Acceptance Length, the mean number of verified tokens generated per forward pass. By structurally aligning the diffusion projections with the exact AR predictive distribution, Orthrus achieves a significantly higher acceptance rate than existing speculative methods. As shown in Figure~\ref{fig:performance_combined} (Left), Orthrus consistently outperforms baselines across reasoning and coding domains. On MATH-500, it reaches an acceptance length of 11.7, substantially surpassing both DFlash (7.9) and EAGLE-3 (3.5).

Furthermore, the architectural efficiency of Orthrus becomes especially pronounced in long-context scenarios, offering several critical advantages over existing methods. First, the shared-cache design introduces absolutely zero penalty to the Time-to-First-Token (TTFT), ensuring highly responsive prefill stages. Second, it effectively mitigates distribution drift as context length scales, maintaining a highly faithful alignment with the base model's target distribution. Most notably, while standard speculative decoding methods like DFlash exhibit a significant deterioration in speed as sequence lengths grow due to the compounding overhead of dual KV caches, Orthrus sustains a near-constant end-to-end throughput. Figure~\ref{fig:performance_combined} (Right) illustrates this stability, demonstrating that Orthrus maintains near-peak generation throughput at context lengths up to 40K, regardless of the underlying sequence size.

\section{Ablation Study}
\paragraph{Effect on Parallel Block Size ($K$).}
\begin{wrapfigure}{r}{0.38\textwidth}
    \vspace{-38pt}
    \centering
    \includegraphics[width=0.85\linewidth]{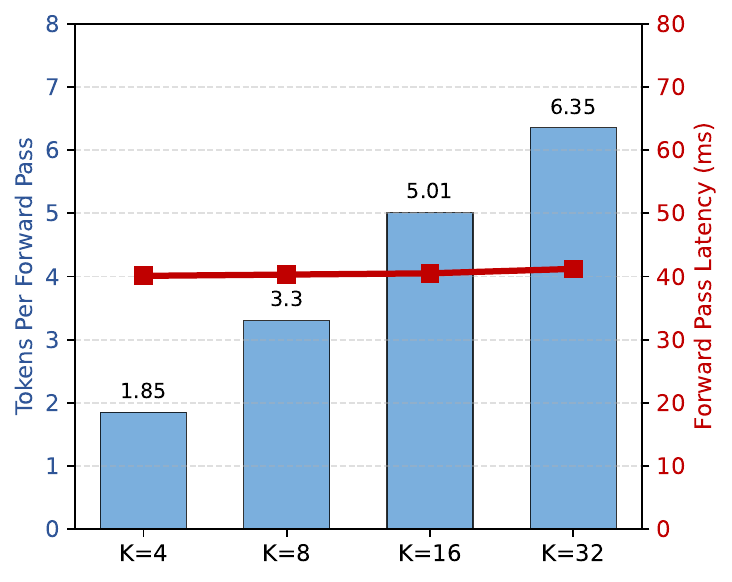}
    \caption{Throughput vs. Latency.}
    \label{fig:ablation_k_plot}
\end{wrapfigure}

We evaluate throughput and latency sensitivity to the parallel block size ($K$) on MATH-500 using Orthrus-Qwen3-8B. By processing the extended block simultaneously against a pre-computed KV cache, the diffusion view maintains a constant forward-pass latency across all evaluated sizes (Figure~\ref{fig:ablation_k_plot}). Scaling to $K=32$ increases the TPF to 6.35, yielding a 3.6$\times$ throughput multiplier over $K=4$ with zero latency penalty. We select $K=32$ as the optimal configuration to maximize parallel acceleration.

\paragraph{Ablation on Multi-Step Denoising.}
\begin{wraptable}{r}{0.38\textwidth}
    \centering
    \small
    \caption{Impact of multi-step denoising.}
    \label{tab:ablation_multistep}
    \begin{tabular}{@{}lcc@{}}
    \toprule
    \textbf{Strategy} & \textbf{Passes} & \textbf{TPF} \\ \midrule
    Multi-Step & 2 & 3.53 \\
    \textbf{Orthrus (Single-Step)} & \textbf{1} & \textbf{6.35} \\ \bottomrule
    \end{tabular}
\end{wraptable}

To validate our single-step projection strategy, we evaluate a multi-step iterative denoising variant adapted from Fast-dLLM-v2 \citep{wu2025fast}. During training, rather than masking all future tokens, we randomly mask 50\% of the block positions and apply a complementary masking strategy across dual views to ensure comprehensive supervision. During inference, it requires two sequential forward passes to fully materialize the block. As detailed in Table~\ref{tab:ablation_multistep}, while iterative refinement is standard in diffusion literature, the additional computational pass degrades throughput. The 2-step prediction strategy slashes the TPF by 1.8$\times$, confirming that single-step projection is optimal for our approach.

\section{Conclusion}
\label{sec:conclusion}

In this work, we introduced \textbf{Orthrus}, a novel dual-architecture framework that fundamentally reconciles the trade-off between autoregressive generation fidelity and diffusion-based parallelism. By embedding a lightweight, trainable diffusion module within a frozen, pre-trained AR backbone, we established a unified system capable of parallel token generation that natively utilizes a shared high-fidelity KV cache. Our empirical results demonstrate that Orthrus effectively breaks the sequential generation bottleneck, delivering up to a $7.8\times$ speedup across diverse mathematical and structural benchmarks while incurring zero redundant memory overhead. By leveraging intra-model consensus, our approach guarantees lossless inference parity, offering a highly efficient and scalable solution for high-throughput deployment. 

\bibliographystyle{plainnat}
\bibliography{references}

\appendix

\section{Training Details}
\label{app:training_details}

To train the Orthrus dual-view architecture, we employ a highly optimized distillation pipeline that isolates the diffusion head while keeping the autoregressive (AR) backbone strictly frozen. Below, we detail the dataset composition, hardware configuration, and hyperparameters utilized to train the models.

\begin{table}[h]
\centering
\caption{\textbf{Training Hyperparameters.} }
\label{tab:hyperparameters}
\begin{tabular}{@{}lc@{}}
\toprule
\textbf{Hyperparameter} & \textbf{Value} \\ \midrule
Max Sequence Length ($L$) & 2048 \\
Anchor Blocks per Sequence & 256 \\
Parallel Block Size ($K$) & 32 \\
Training Epochs & 2 \\
Peak Learning Rate & $2 \times 10^{-4}$ \\
Learning Rate Scheduler & Cosine \\
Warmup Ratio & 0.05 \\
Gradient Clipping & 1.0 \\
Micro Batch Size & 1 \\
Gradient Accumulation Steps & 16 \\
Global Batch Size & 128 \\
Precision & \texttt{bfloat16} \\
Data Domain Split & 1:1:1 (Chat:Math:Code) \\ \bottomrule
\end{tabular}
\end{table}

\paragraph{Datasets.}
To ensure robust performance across diverse domains, the training corpus is constructed by sampling from the open-source Nemotron-Post-Training-Dataset-v2 \citep{NemotronPostTrainingDatasetV2}. To guarantee exact distributional alignment, target outputs are generated directly by the frozen AR head and used as the distillation signal to train the diffusion head. The sampled data is strictly balanced across three domains: \textbf{Mathematical Reasoning}, \textbf{Code Generation}, and \textbf{General Chat \& Instruction Tuning}.
During data loading, we enforce a uniform sampling strategy across these three categories. To maximize hardware utilization, we employ sequence packing, concatenating supervised examples up to a strict maximum sequence length, denoted as $L = 2048$ tokens. This packing strategy yields $471,952$ training instances, equivalent to $0.96$B total tokens. For each packed sequence, we uniformly sample exactly 256 anchor blocks. The diffusion view utilizes the \texttt{<mask>} tokens within these blocks to predict the corresponding future token trajectory by minimizing the forward KL divergence against the frozen AR teacher's exact predictive distribution. Only the parameters of the newly injected diffusion attention heads are updated.

\paragraph{Hyperparameters.}
Training was executed on a single computational node equipped with 8 GPUs (e.g., 8$\times$H200). We utilize PyTorch FSDP-2 with a micro-batch size of 1 per device and 16 gradient accumulation steps, yielding an effective global batch size of 128 sequences per optimization step. The diffusion parameters are optimized in \texttt{bfloat16} precision to reduce memory overhead and accelerate computation. The model is trained for 2 epochs using a cosine learning rate scheduler with a peak learning rate of $2 \times 10^{-4}$ and a 5\% warmup ratio. To ensure stability during training, we apply gradient clipping with a maximum norm of 1.0. A summary of the training hyperparameters is provided in Table~\ref{tab:hyperparameters}.

\section{Analysis}
\label{app:analysis}

\paragraph{Training Objective. }
\begin{wraptable}{r}{0.48\textwidth}
    \centering
    \vspace{-12pt}
    \small
    \caption{Impact of the Training Objective.}
    \label{tab:ablation_objective}
    \begin{tabular}{@{}lcc@{}}
    \toprule
    \textbf{Objective} & \textbf{Accuracy} & \textbf{TPF} \\ \midrule
    Cross-Entropy (Hard) & 86.2\% & 5.86 \\
    \textbf{KL Divergence (Soft)} & \textbf{86.2\%} & \textbf{6.35} \\ \bottomrule
    \end{tabular}
    \vspace{-10pt}
\end{wraptable}
In our standard configuration, the diffusion view is trained via forward KL divergence (Equation \ref{eq:kl_div}) to distill the full predictive distribution of the AR teacher. To validate this design, we ablate our soft distillation objective by training a variant of Orthrus-Qwen3-8B using standard Cross-Entropy (CE) against the hard ground-truth tokens of the dataset.

As shown in Table~\ref{tab:ablation_objective}, while the exact intra-model consensus mechanism guarantees that both models achieve the identical baseline accuracy on MATH-500, their inference speeds diverge dramatically. Training with hard labels causes the diffusion head to overfit to the dataset's surface syntax rather than internalizing the specific causal trajectory preferred by the AR base model. During inference, this structural misalignment triggers high rejection rates in the consensus validation phase, slashing the Effective Tokens Per Forward Pass (TPF) from 6.35 down to 5.86. 

\begin{figure}[htbp]
    \centering
    \includegraphics[width=\textwidth]{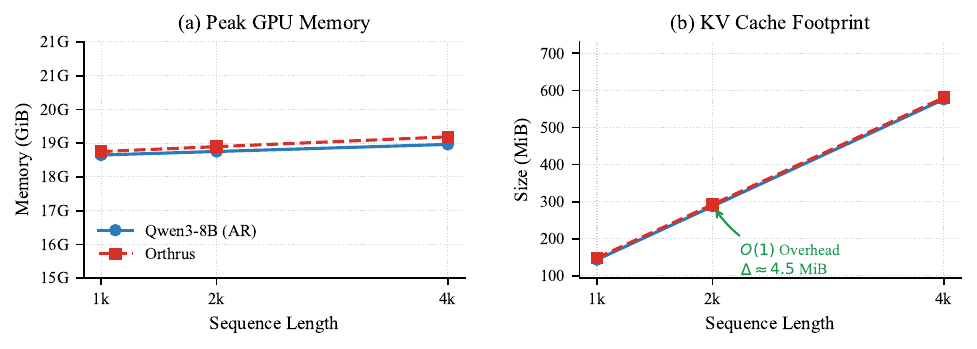}
    \caption{\textbf{Memory footprint scaling of Orthrus versus the Qwen3-8B baseline.} \textbf{(a)} The peak GPU memory overhead is practically negligible ($<1\%$), demonstrating that the dual-view architecture minimizes VRAM penalties. \textbf{(b)} The KV cache footprint exhibits a strictly constant $O(1)$ overhead ($\approx 4.5$ MiB) across all sequence lengths. By completely sharing the historical AR cache, Orthrus natively bypasses the linear cache redundancy typical of standard speculative decoding.}
    \label{fig:memory_scaling}
\end{figure}

\paragraph{Memory Footprint Scaling.}
A key advantage of the Orthrus architecture is its extreme memory efficiency during inference, particularly regarding peak GPU memory and dynamic Key-Value (KV) cache. By integrating the diffusion view alongside the standard autoregressive attention mechanism, Orthrus maintains a lean memory profile. Across varying sequence lengths, the peak GPU memory penalty is negligible ($\sim 100$ MiB). Furthermore, the architecture eliminates the redundant KV cache overhead typical of multi-model speculative decoding. Because the diffusion head conditions directly on the AR head's causal cache, the only additional memory required is the transient state for the fixed-size parallel projection block ($K=32$). Consequently, Orthrus exhibits a strictly constant $O(1)$ KV cache overhead. As demonstrated in Figure~\ref{fig:memory_scaling}, this manifests as a fixed $\Delta \approx 4.5$ MiB increase regardless of the total sequence length $L$, allowing the framework to scale to massive context windows without compounding memory degradation.

\section{Limitation}
\label{app:limitation}

Because the Orthrus architecture strictly freezes the autoregressive backbone to guarantee exact inference parity, its generative capabilities are strictly upper-bounded by the foundation model. The diffusion head is distilled exclusively to mirror the AR teacher's exact predictive distribution. Consequently, the framework acts solely as an inference accelerator and inherently inherits any biases, knowledge gaps, or hallucination tendencies present in the underlying base model.

\end{document}